\pdfoutput=1
\documentclass[11pt]{article}

\usepackage[]{acl2021}

\usepackage{times}
\usepackage{latexsym}

\usepackage[T1]{fontenc}
\usepackage[utf8]{inputenc}

\usepackage{graphicx}
\usepackage{colortbl}
\usepackage{amsfonts}
\usepackage[normalem]{ulem}
\useunder{\uline}{\ul}{}
\usepackage{amssymb}
\usepackage{amsfonts}
\usepackage{framed}
\usepackage{color}
\usepackage{multicol}
\usepackage{multirow}
\usepackage{makecell}
\usepackage{amsmath}
\usepackage{booktabs}
\usepackage{array}
\usepackage{adjustbox}
\usepackage{arydshln}
\usepackage{paralist}
\usepackage{algorithm}
\usepackage{bbm}
\usepackage[figuresright]{rotating}
\usepackage[noend]{algpseudocode}
\allowdisplaybreaks[4]
\usepackage{microtype}
\usepackage{tikz}
\newcommand*{\circled}[1]{\lower.7ex\hbox{\tikz\draw (0pt, 0pt)%
    circle (.5em) node {\makebox[1em][c]{\small #1}};}}

\title{On the Robustness of Question Rewriting Systems \\ to Questions of Varying Hardness}

\author{Hai Ye\textsuperscript{\rm 1} \ \ \ Hwee Tou Ng\textsuperscript{\rm 1} \ \ \ Wenjuan Han\textsuperscript{\rm 2} \\ 
\textsuperscript{\rm 1}Department of Computer Science, National University of Singapore \\
\textsuperscript{\rm 2}Beijing Institute for General Artificial Intelligence (BIGAI), Beijing, China \\
\texttt{\{yeh,nght\}}\texttt{@comp.nus.edu.sg} \\
\texttt{hanwenjuan@bigai.ai}
}

\date{}

\begin{document}
\maketitle
\begin{abstract}

In conversational question answering (CQA), the task of question rewriting~(QR) in context aims to rewrite a context-dependent question into an equivalent self-contained question that gives the same answer. In this paper, we are interested in the robustness of a QR system to questions varying in rewriting hardness or difficulty. Since there is a lack of questions classified based on their rewriting hardness, we first propose a heuristic method to automatically classify questions into subsets of varying hardness, by measuring the discrepancy between a question and its rewrite. To find out what makes questions hard or easy for rewriting, we then conduct a human evaluation to annotate the rewriting hardness of questions. Finally, to enhance the robustness of QR systems to questions of varying hardness, we propose a novel learning framework for QR that first trains a QR model independently on each subset of questions of a certain level of hardness, then combines these QR models as one joint model for inference. Experimental results on two datasets show that our framework improves the overall performance compared to the baselines\footnote{Our source code is available at~\url{https://github.com/nusnlp/DiffQRe}. This work was done while Wenjuan Han was a research fellow at the National University of Singapore.}. 

\end{abstract}

\section{Introduction}
In conversational question answering (CQA) \cite{DBLP:conf/emnlp/ChoiHIYYCLZ18,DBLP:journals/tacl/ReddyCM19}, several sequential questions need to be answered one by one given a relevant article. To answer a question in CQA, we need to understand the historical context of the question. For example, to answer the question ``\emph{When did he begin writing these pieces?}'', we need to know what \emph{he} refers to in the conversation context. In our work, we address question-in-context rewriting~(QR), which aims to rewrite a context-dependent question into an equivalent self-contained question in CQA, e.g., replacing \emph{he} in the above example with its referent from the context. The task is formulated as a text generation task that generates the rewrite of a question given the original question and its conversation context~\cite{DBLP:conf/emnlp/ElgoharyPB19}.

% \usepackage{graphicx}
\begin{table}[]
\centering
\small
{%
\begin{tabular}{p{7cm}}
 \toprule[.8pt]
\textbf{Topic words}: Benigno Aquino III; Senate (2007 - 10)              \\ \hdashline[3pt/5pt]
$\mathbf{q_1}$: What changes did he make while in the Senate?                \\
$\mathbf{a_1}$: I don't know.                                                \\ \hdashline[3pt/5pt]
$\mathbf{q_2}$: When was {\emph{he}} elected?                                         \\
$\rightarrow \mathbf{q'_2}$: When was {\emph{Benigno Aquino III}} elected to Senate?             \\
$\mathbf{a_2}$: May 15, 2007                                                \\ \hdashline[3pt/5pt]
$\mathbf{q_3}$: Was he a republican or democrat?                            \\
$\mathbf{a_3}$: Genuine Opposition (GO), a coalition comprising a number of parties, including Aquino's own Liberal Party, ...                           \\ \hdashline[3pt/5pt]
$\mathbf{q_4}$: Are there any other interesting aspects about this article? \\
$\rightarrow \mathbf{q'_4}$: Are there any other interesting aspects about Benigno Aquino III article {\emph{aside from political affiliation or when Benigno was elected?}} \\
$\mathbf{a_4}$: Aquino was endorsed by the pentecostal Jesus Is Lord Church. \\
\bottomrule[0.8pt]
\end{tabular}
\caption{One dialogue example from \citet{DBLP:conf/emnlp/ElgoharyPB19} including questions~($\mathbf{q_i}$) and answers~($\mathbf{a_i}$) and certain rewrites~($\mathbf{q'_i}$) of the questions.}\label{tab:example}
}
\end{table}

We are interested in how robust a QR system is to questions with different rewriting hardness~(or difficulty). As we can see from the examples in Table~\ref{tab:example}, rewriting the question $\mathbf{q_2}$ requires only replacing the pronoun \emph{he} by its referent, which usually appears in the conversation context, and the model can identify the referent by \emph{attention} \cite{DBLP:conf/emnlp/LuongPM15}. However, for the question $\mathbf{q_4}$, to find the missing \emph{aside from} clause, the model needs to understand the entire conversation since the question asks about other interesting aspects about the article related to the topic of the entire conversation. Understanding the whole context will be challenging for the model. Can a QR model still work well when rewriting the hard questions? 

In section \ref{sec:robustness-eval}, our first study is on evaluating the performance of a QR model under questions varying in hardness. One issue in this process is that there is a lack of classified questions in different rewriting hardness. Though we can rely on human labor to annotate the questions, it is expensive and not scalable. Instead, we propose a simple yet effective heuristic method to classify the questions automatically. We measure the \emph{discrepancy} between a question and its rewrite, where the larger the discrepancy, the more difficult to rewrite the question. The intuition is that if a question is very dissimilar to its rewrite, more information has to be filled into the rewrite, which means the question is harder to rewrite. We specifically use the BLEU score to measure the discrepancy, and lower scores mean larger discrepancies. Using this method, we then split the questions into three subsets: hard, medium, and easy, and evaluate the baseline systems using these subsets. 

In order to verify the classified subsets and find out what makes questions different in rewriting difficulty, in section \ref{sec:human-eval}, we further evaluate the question characteristics in hard, medium, and easy subsets through human evaluation. We first manually summarize the commonly used rules for rewriting questions from the training set, and then annotate the questions using the labels of summarized rewriting rules, followed by counting the number of these rewriting rules used in these subsets. 

Finally, to enhance the robustness of a QR model to questions varying in difficulties, we propose a novel learning framework in section \ref{sec:model}, where we first separately train a QR model on each hard, medium, and easy subset, and then combine these models into a joint model for inference. Training one sole model on each subset is to let the model better learn domain-specific information to deal with one specific type of questions~(hard/medium/easy). By combining the models together, we have a joint model capable of rewriting questions differing in rewriting hardness. Specially, we introduce adapters~\cite{DBLP:conf/icml/HoulsbyGJMLGAG19} to reduce parameters when building private models and we present sequence-level adapter fusion and distillation~(SLAF and SLAD) to effectively combine the private models into a joint model.

Our contributions in this paper include: 
\begin{compactitem}
    \item We are the {\em first} to study the robustness of a QR system to questions with varying levels of rewriting hardness;
    \item We propose an effective method to identify questions of different rewriting hardness;
    \item We manually annotate questions sampled from the subsets with summarized rewriting rules for validity and address what makes questions hard or easy for rewriting;
    \item We propose a novel QR framework by taking into account the rewriting hardness.
\end{compactitem}

We have the following observations in our paper:
\begin{compactitem}
    \item The baseline systems perform much worse on the hard subset but perform well on the easy subset;
    \item We find that easy questions usually only require \emph{replacing pronouns} but hard questions involve more complex operations like \emph{expanding special Wh* questions};
    \item Experiments show that our QR learning framework enhances the rewriting performance compared to the baselines.  
    
\end{compactitem}

\section{Related Work}

\citet{DBLP:conf/emnlp/ElgoharyPB19} created the QR dataset which rewrites a subset of the questions from QuAC~\cite{DBLP:conf/emnlp/ChoiHIYYCLZ18}. Based on this dataset, some recent work has studied this task and formulates QR as a text generation task with an encoder-decoder architecture~\cite{DBLP:conf/emnlp/ElgoharyPB19,DBLP:conf/coling/KumarJ16,DBLP:journals/corr/abs-2004-14652,DBLP:conf/acl-mrqa/LiSNWLFZ19,DBLP:journals/corr/abs-2004-01909}. 

The difficulty of answering a question given a relevant document has been studied in the question answering community~\cite{DBLP:conf/naacl/DuaWDSS019,DBLP:journals/tacl/WolfsonGGGGDB20}. \citet{DBLP:conf/emnlp/SugawaraISA18} examine 12 reading comprehension datasets and determine what makes a question more easily answered. \citet{DBLP:conf/emnlp/PerezLYCK20,DBLP:conf/acl/MinZZH19,DBLP:conf/naacl/TalmorB18,DBLP:conf/emnlp/DongMRL17} study how to make a hard question more easily answered. However, there is no work to date that studies whether rewriting difficulties exist in QR and how to measure the difficulties. Some other work is similar to QR but focuses on other tasks such as dialogue tracking~\cite{DBLP:conf/naacl/RastogiGCM19,DBLP:conf/acl/SuSZSHNZ19,DBLP:conf/emnlp/LiuCLZZ20} and information retrieval~\cite{DBLP:conf/sigir/VoskaridesLRKR20,DBLP:journals/corr/abs-2005-02230,DBLP:conf/cikm/LiuZYCY19}.

Varying rewriting difficulties can result in multiple underlying data distributions in the QR training data. The shared-private framework has been studied to learn from training data with multiple distributions~\cite{DBLP:conf/naacl/ZhangDS18,DBLP:conf/acl/LiuQH17}. 
One issue of the shared-private framework is parameter inefficiency when building private models
We use adapter tuning~\cite{DBLP:conf/cvpr/RebuffiBV18,DBLP:conf/nips/RebuffiBV17} to build the private models. Adapter tuning was recently proposed for adapting a pre-trained language model, e.g., BERT~\cite{DBLP:conf/naacl/DevlinCLT19}, to downstream tasks~\cite{DBLP:journals/corr/abs-2005-00247,DBLP:conf/emnlp/PfeifferVGR20,DBLP:conf/icml/HoulsbyGJMLGAG19}, and its effectiveness has been verified by previous work~\cite{DBLP:conf/emnlp/BapnaF19,DBLP:conf/emnlp/PfeifferRPKVRCG20,DBLP:journals/corr/abs-2002-01808,DBLP:conf/acl/HeLYTDCLBS20}. We are the first to apply it to reduce model parameters in the shared-private framework. How to combine the knowledge stored in multiple adapters is also important. \citet{DBLP:journals/corr/abs-2005-00247} propose adapter fusion to build an ensemble of adapters in multi-task learning. We propose sequence-level adapter fusion in our work.

\section{Question-in-Context Rewriting}
Question-in-context rewriting~(QR) aims to generate a self-contained rewrite from a context-dependent question in CQA. Given a conversational dialogue $\mathcal{H}$ with sequential question and answer pairs $\{\mathbf{q}_1, \mathbf{a}_1, \cdots, \mathbf{q}_n, \mathbf{a}_n\}$, for a question $\mathbf{q}_i$ from $\mathcal{H}$ with its history $\mathbf{h}_i = \{\mathbf{q}_1, \mathbf{a}_1, \cdots, \mathbf{q}_{i-1}, \mathbf{a}_{i-1}\}$, we generate its rewrite $\mathbf{q}'_i$. 
We define the labeled dataset $\mathcal{D} = \{\mathbf{q}_i, \mathbf{h}_i, \mathbf{q}'_i\}_{i=1}^{|\mathcal{D}|}$ which is a set of tuples of question $\mathbf{q}$, history $\mathbf{h}$, and rewrite $\mathbf{q}'$. Following previous work~\cite{DBLP:conf/emnlp/ElgoharyPB19}, we model QR in an encoder-decoder framework, by estimating the parameterized conditional distribution for the output $\mathbf{q}'$ given the input question $\mathbf{q}$ and history $\mathbf{h}$. For $(\mathbf{q},\mathbf{h},\mathbf{q}') \in \mathcal{D}$, we minimize the following loss function parameterized by $\theta$:
\setlength\abovedisplayskip{4pt}\setlength\belowdisplayskip{-4pt}
\begin{flalign}
 &\mathcal{L}^{\theta}_{NLL} = -\log P(\mathbf{q}'|\mathbf{q},\mathbf{h};\theta) \nonumber \\
&= -\sum_{t=1}^{T_{\mathbf{q}'}}\sum_{k=1}^{|V|}\mathbbm{1}\{q'_t=k\}\log P(q'_t = k|\mathbf{q}'_{<t}, \mathbf{q}, \mathbf{h};\theta)
\label{eq:nll}
\end{flalign}
in which $T_{\mathbf{q}'}$ is the length of $\mathbf{q}'$ and $|V|$ is the vocabulary size. 
Following \citet{DBLP:conf/emnlp/ElgoharyPB19}, $\mathbf{q}$ and $\mathbf{h}$ are concatenated into one sequence as the input. All previous turns of the history information are combined for learning. The choice of the encoder-decoder framework can be LSTM~\cite{DBLP:conf/emnlp/ElgoharyPB19}, transformer~\cite{DBLP:journals/corr/abs-2004-14652}, or pre-trained language models~\cite{DBLP:journals/corr/abs-2004-01909}. In our work, we build our model based on the pre-trained language model BART~\cite{DBLP:conf/acl/LewisLGGMLSZ20}. 

% \usepackage{graphicx}

\section{Difficulty of Question Rewriting}

The difficulty of rewriting a question varies across questions. We propose a simple yet effective heuristic to formulate rewriting difficulty as the discrepancy between a question and its rewrite. To generate a self-contained rewrite, we need to identify relevant information from the conversation context to incorporate it into the original question. We observe that if the discrepancy is large, we need to identify more missing information from the conversation context which makes the rewriting task more difficult. 

In this work, we use BLEU score to measure the discrepancy.  BLEU has been widely used to measure how similar two sentences are~\cite{DBLP:conf/acl/PapineniRWZ02}. Given a question $\mathbf{q}$ and its rewrite $\mathbf{q}'$, we define the difficulty score $\mathrm{z}$ for rewriting $\mathbf{q}$ as:
\begin{equation}
\setlength\abovedisplayskip{4pt}\setlength\belowdisplayskip{4pt}
    \mathrm{z} = BLEU(\mathbf{q}, \mathbf{q}')
\end{equation}
where the rewrite $\mathbf{q}'$ is the reference and $\mathrm{z} \in [0,1]$. A low $\mathrm{z}$ score indicates a larger discrepancy between $\mathbf{q}$ and $\mathbf{q}'$, making it more difficult to rewrite $\mathbf{q}$ into $\mathbf{q}'$. Besides BLEU, we also study the effectiveness of ROUGE, lengths of $\mathbf{q}$ and $\mathbf{q}'$, and $|\mathbf{q}|/|\mathbf{q}'|$ in $\S$\ref{sec:FA} to measure rewriting difficulty.

\section{Difficulty-Aware QR with Adapters}\label{sec:model}

Previous work on QR learns to rewrite questions with only one shared model~\cite{DBLP:conf/emnlp/ElgoharyPB19}, which cannot adequately model all questions with different rewriting difficulties. Instead of using only one shared model, we propose a novel method to classify a question into several classes by measuring its rewriting difficulty~($\S$\ref{sec:QC}), learn a private model for each class~($\S$\ref{sec:LPM}), and finally combine the private models for inference~($\S$\ref{sec:ME}). Different questions with varying rewriting difficulties result in multiple data distributions in the training set. By dividing the training data into several classes with varying rewriting difficulties, we can better learn the data distributions with the help of private models~\cite{DBLP:conf/naacl/ZhangDS18}.

\subsection{Question Classification}\label{sec:QC}

We compute the difficulty score $\mathrm{z}$ of each question in the dataset. We set score intervals and group the questions with difficulty scores within the same interval together. Specifically, we divide the original dataset $\mathcal{D}$ into $m$ classes: $\{\mathcal{D}_1, \mathcal{D}_2, \cdots, \mathcal{D}_m \}$. Setting $m$ to a large number (e.g., the number of training samples) can more accurately model the data distribution of the training data, but at the expense of data sparsity in each class such that a private model cannot be adequately trained.

\begin{figure}[t]
\begin{center}
\includegraphics[width=\columnwidth]{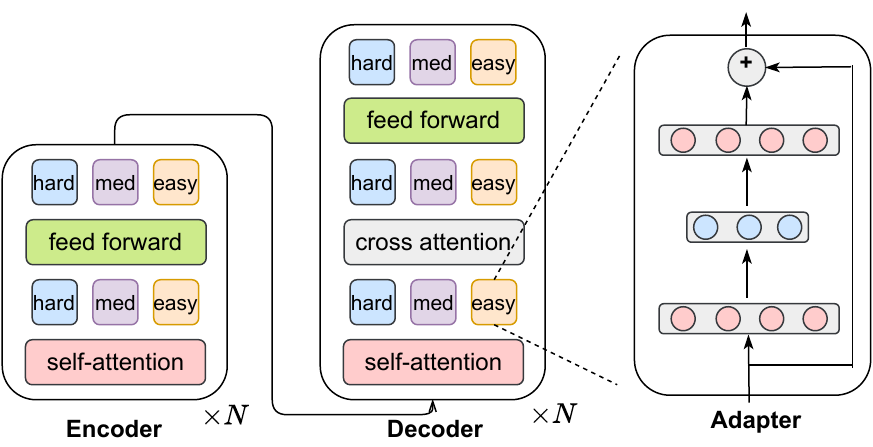}
\end{center}
\caption{Illustration of our model architecture. Class-private adapters are added into the transformer. The original PLM weights are shared across all private models. $N$ is the number of transformers in the encoder and decoder.}
\label{fig:model}
\end{figure}

\subsection{Learning Private Models}\label{sec:LPM}
After dividing the questions into $m$ classes, we learn a private model for each class. By training on each class of data, the private model can better learn the domain-specific information. The common way to use a pre-trained language model~(PLM) such as BART is to fine-tune the model on the downstream task. However, doing so will require $m$ times the number of PLM parameters to build all private models, where $m$ is the number of classes. This results in a large number of parameters, leading to inefficiency.

To reduce the number of model parameters in learning the private models, we introduce adapters into the PLM. Adapters are light-weight neural networks and are plugged into the PLM. When adapting the PLM to downstream tasks, we only need to update the parameters of the adapters but keep the original parameters of the PLM frozen and shared among all private models. Where to place the adapters in the neural architecture  will affect the efficacy of adapters. As shown in Figure~\ref{fig:model}, for each transformer layer in the encoder, we add the adapters after the self-attention layer and feed-forward layer. We further add the adapters after the cross-attention layer in the decoder. Though our model is built on BART, our proposed placement of adapters can also be used in other PLMs, such as T5~\cite{DBLP:journals/jmlr/RaffelSRLNMZLL20}.

In Figure~\ref{fig:model}, the adapter is a module with a stack of two linear layers following \citet{DBLP:conf/icml/HoulsbyGJMLGAG19}. Formally, given an input hidden vector $\mathbf{x}$ from the previous layer, we compute the output hidden vector $\mathbf{x}'$ of the adapter as:
\begin{equation}
\setlength\abovedisplayskip{4pt}\setlength\belowdisplayskip{4pt}
    \mathbf{x'} =  f_2(\tanh(f_1(\mathbf{x}))) + \mathbf{x}
\end{equation}
where $f_1(\cdot)$ is the down-scale linear layer and $f_2(\cdot)$ is the up-scale linear layer. The hidden vector size is smaller than the dimension of the input vector. Learning a private model for one class only introduces $5 \times N$ adapters, where $N$ is the number of layers in the encoder and decoder. The original parameters of the PLM are shared by all adapters, so the number of parameters required when building the private models can be much reduced.

\begin{figure}[t]
\begin{center}
\includegraphics[width=\columnwidth]{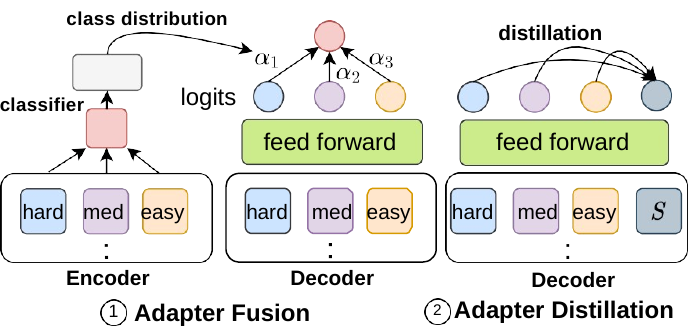}
\end{center}
\caption{Illustration of our sequence-level adapter fusion and distillation.}
\label{fig:ensemble}
\end{figure}

\subsection{Model Ensemble}\label{sec:ME}

After learning the private models for all classes, at test time, we present the question to the corresponding private model to generate its rewrite if we know which class this question belongs to. However, it is not possible to determine the difficulty score by calculating the BLEU score between the question and its rewrite since there is no gold-standard rewrite for the question at test time. As such, we need to combine the private models into one model for inference. In this work, we propose two methods to combine the private models, as explained below.

\noindent{\textbf{Sequence-level Adapter Fusion~(SLAF).}} \ \ After dividing the training set into $m$ classes based on the difficulty scores, we assign a difficulty label to each class to obtain a set of class labels $\{l_1, l_2, \cdots, l_m\}$. We introduce a classifier to learn to predict the difficulty label $l$, given a question $\mathbf{q}$ and its conversation history $\mathbf{h}$. As shown in Figure~\ref{fig:ensemble}, during inference, we obtain the logistic output from each private model. The classifier generates the class distribution to combine the logistic outputs for sequence generation. 

By assigning a difficulty label to each question, we obtain the dataset $\mathcal{D}' = \{\mathbf{q}_i, \mathbf{h}_i, \mathbf{q}'_i, l_i\}_{i=1}^{|\mathcal{D}'|}$. For each training sample $(\mathbf{q}, \mathbf{h}, \mathbf{q}', l) \in \mathcal{D}'$, we minimize the following loss function:
\begin{equation}
\setlength\abovedisplayskip{4pt}\setlength\belowdisplayskip{4pt}
\begin{aligned}
    \mathcal{L}^{\theta_{c}}_{NLL} = -\log \mathrm{softmax}\big (\sum_{i=1}^m \alpha_i f_i(\mathbf{q},\mathbf{h}; \theta_i) \big ) \\
    - \log P(l|\mathbf{q}, \mathbf{h}; \theta_c)
\end{aligned}\label{eq:alpha}
\end{equation}
where $f_i$ is the $i$th private model, $\alpha_i$ is the class weight of the $i$th private model, and $\theta_c$ is the parameter of the classifier. We jointly estimate the conditional distribution for sequence generation and the distribution for classification. In this process, the private models are frozen and not updated. We combine the vectors out of the private models to calculate the vector $f_c$ as the input for the classifier: 
\begin{equation}
\setlength\abovedisplayskip{4pt}\setlength\belowdisplayskip{0pt}
    f_c = \frac{1}{m} \sum_{i=1}^{m} f^i_{encoder}(\mathbf{q}, \mathbf{h}; \theta_i)
\end{equation}
where $f^i_{encoder}$ is the encoder of the $i$th private model. For each private model, we average the token embeddings from the last layer of the encoder.

\noindent{\textbf{Sequence-level Adapter Distillation~(SLAD).}} \ \ SLAF provides a way to combine the private models, but it is time-consuming during inference since it requires each private model to compute its logistic output before combination. Another drawback is that the domain classifier in SLAF cannot generate the best class distributions at test time, causing non-optimal rewriting results by SLAF.  
As shown in Figure~\ref{fig:ensemble}, to speed up inference and better combine the private models, we distill the private models into one shared model. We expect the student model $S$~(modeled by adapters) to be able to generate questions with different rewriting difficulties. For each training sample $(\mathbf{q}, \mathbf{h}, \mathbf{q}', l) \in \mathcal{D}'$, we define the knowledge distillation loss function as follows:
\begin{equation}
\setlength\abovedisplayskip{0pt}\setlength\belowdisplayskip{-2pt}
\begin{aligned}
       \mathcal{L}^{\theta_S}_{KD} = -\sum_{t=1}^{T_{\mathbf{q}'}}\sum_{k=1}^{|V|}P^{(l)}\{{q}'_t=k|\mathbf{q}'_{<t},\mathbf{q},\mathbf{h};\theta^{(l)}\} \\
    \times \log P({q}'_t = k|\mathbf{q}'_{<t},\mathbf{q},\mathbf{h};\theta_S)
\end{aligned}
\end{equation}
in which we approximate the output distribution of the teacher private model $l$ parameterized by $\theta^{(l)}$ with the student model parameterized by $\theta_S$. We learn the student model with the following function:
\begin{equation}
\setlength\abovedisplayskip{4pt}\setlength\belowdisplayskip{4pt}
    \mathcal{L}^{\theta_S}_{distill} = (1 - \gamma) \cdot \mathcal{L}^{\theta_S}_{KD} + \gamma \cdot \mathcal{L}^{\theta_S}_{NLL}
    \label{eq:distill}
\end{equation}
where $\mathcal{L}^{\theta_S}_{NLL}$ is the same loss function in Eq.~\ref{eq:nll}, and $\gamma$ is a hyper-parameter. The private models are fixed in the distillation process. Since we directly distill the knowledge of the private models into a shared model without the soft weights generated by the domain classifier from SLAF, SLAD can better combine the private models and achieve better rewriting performance.

\begin{table}[]
\centering
\small
\resizebox{7cm}{!}{%
\begin{tabular}{lcccc}
\hline
                & \textbf{{Train}}  & \textbf{Valid} & \textbf{Test}   & \textbf{All}    \\ \hline
\textsc{canard} & 31,526 & 3,430 & 5,571  & 40,527 \\ 
\textsc{QReCC} & 57,150 & 6,351 & 16,451 & 79,952 \\ \hline
\end{tabular}%
}
\caption{Data splits of \textsc{canard} and \textsc{QReCC}.}
\label{tab:dataset}
\end{table}

\begin{table}[]
\centering
\resizebox{\columnwidth}{!}{%
\begin{tabular}{lcccc}
\toprule[.8pt]
\textsc{canard}          & \textbf{Hard}     & \textbf{Medium}     & \textbf{Easy}  & \textbf{All}    \\ \hline
Ratio (\%) & 32.36    & 33.45      & 34.20   & -  \\
BLEU score    & [0, 0.2) & [0.2, 0.5) & [0.5, 1] & - \\ 
Avg. \# tokens in $\mathrm{q}$ + $\mathrm{h}$ & 111.98 & 103.53 & 90.23  & 101.72\\
Avg. \# tokens in $\mathrm{q}'$ & 14.46 & 11.46 & 9.95  & 11.60\\ \hline \hline
\textsc{QReCC}          & \textbf{Hard}     & \textbf{Medium}     & \textbf{Easy}  & \textbf{All}    \\ \hline

Ratio (\%) & 29.53    & 41.20      & 29.27   & -  \\
BLEU score    & [0, 0.2) & [0.2, 0.4) & [0.4, 1] & - \\ 
Avg. \# tokens in $\mathrm{q}$ + $\mathrm{h}$ & 126.07 & 106.92 & 94.53  & 108.95\\
Avg. \# tokens in $\mathrm{q}'$ & 14.72 & 10.36 & 10.07  & 11.56 \\ \bottomrule[.8pt]
\end{tabular}%
}
\caption{Statistics of each class for the training set of \textsc{canard} and \textsc{QReCC}.} 
\label{tab:data_dis_3_types}
\end{table}

\section{Experiments}

\subsection{Dataset}

We conduct our experiments on \textsc{canard}~\cite{DBLP:conf/emnlp/ElgoharyPB19} and \textsc{QReCC}~\cite{DBLP:conf/naacl/AnanthaVTLPC21}, which are designed for the task of question rewriting in CQA. \textsc{canard} was created from QuAC~\cite{DBLP:conf/emnlp/ChoiHIYYCLZ18}, by rewriting a subset of the questions by humans. The dataset consists of tuples of question, conversation history, and rewrite. \textsc{QReCC} answers conversational questions within large-scale web pages. Detailed data splits for the two datasets are shown in Table~\ref{tab:dataset}. We divide the questions into hard, medium, and easy classes, and the statistics are presented in Table~\ref{tab:data_dis_3_types}.

% \usepackage{graphicx}
\begin{table}[]
\centering
\small
\resizebox{7cm}{!}{%
\begin{tabular}{lcccc}
\toprule[0.8pt]
Model $\diagdown$ $\mathcal{D}$   & \textbf{Hard}              & \textbf{Medium}            & \textbf{Easy}    & \textbf{Mean}          \\ \hline
LSTM-S      & 26.29 & 50.79 & 79.41  & 49.81  \\ 
Fine-tune-S & 39.38             & 53.70             & 66.33  &   53.14        \\ 
Adapter-S   & 39.20             & 53.14             & 65.97  &    52.77       \\ \bottomrule[.8pt]
\end{tabular}%
}
\caption{BLEU scores (in \%) on hard, medium, and easy classes from \textsc{canard}, based on the shared model. Fine-tune-S and Adapter-S are based on BART-base.}
\label{tab:result_3_domains}
\end{table}

\begin{figure}[t]
\begin{center}
\includegraphics[width=\columnwidth]{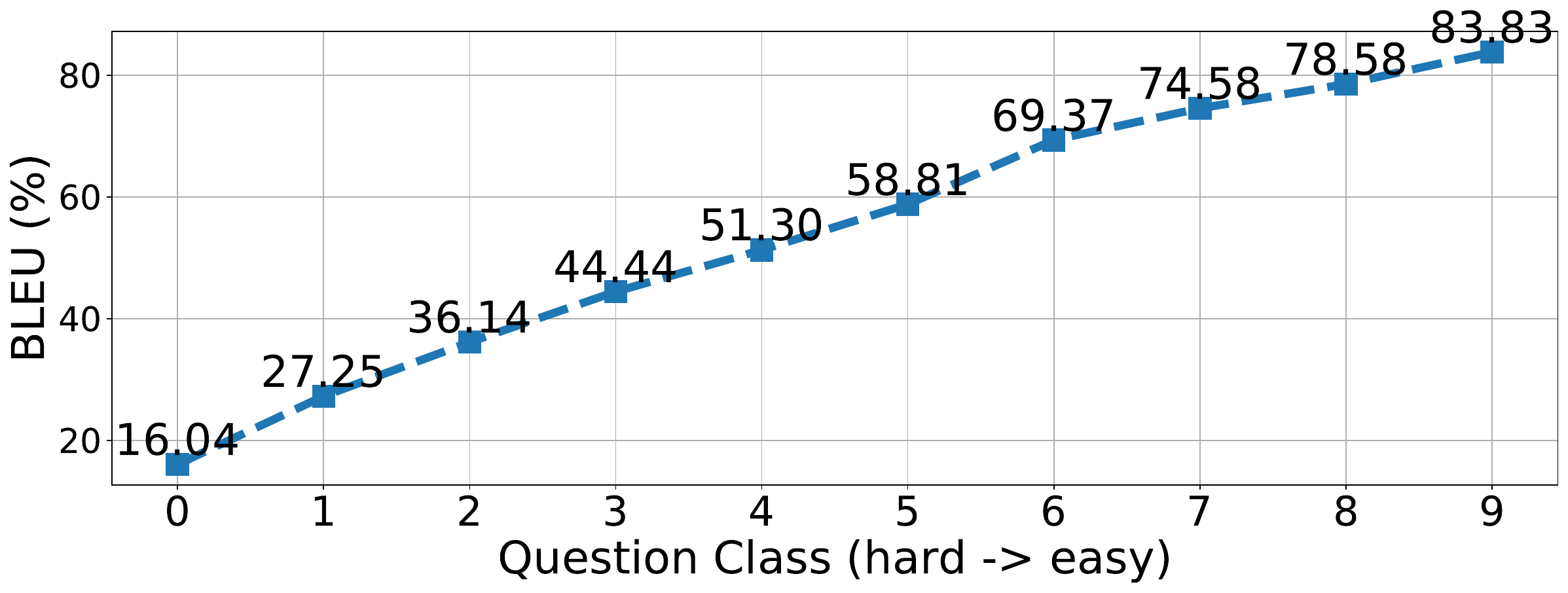}
\end{center}
\caption{10-class BLEU scores on \textsc{canard} with LSTM-S.}
\label{fig:10_dis}
\end{figure}

\begin{figure}[t]
\begin{center}
\includegraphics[width=\columnwidth]{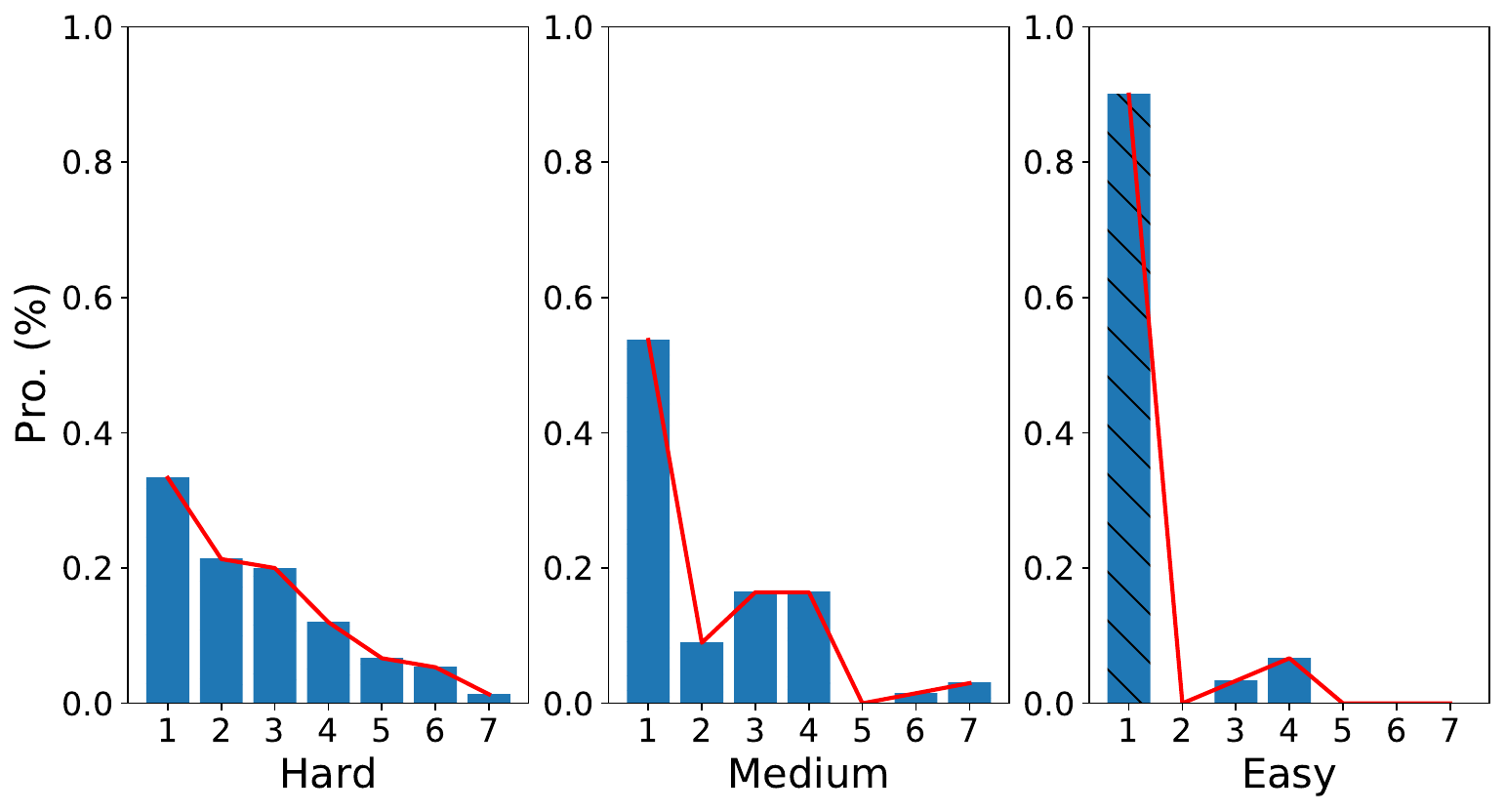}
\end{center}
\caption{The distributions of rewriting rules on hard, medium, and easy subsets in \textsc{canard}.}
\label{fig:3_type_dis}
\end{figure}

% \usepackage{graphicx}
\begin{table*}[]
\centering
\small
%\tiny
%\footnotesize
%\scriptsize
\resizebox{\textwidth}{!}{%
\begin{tabular}{lp{3.5cm}p{9.5cm}}
\toprule
  & \textbf{Rewriting Rules}                                                              & \textbf{Examples}                                                                                                      \\ \hline
1 & replace pronoun, e.g., he/his/she/her/they/their/it/its...                    & when was \textbf{\emph{he}} born ? $\rightarrow$ when was \textbf{\emph{Corbin Bleu}} born ?                                                       \\ \hdashline[4pt/5pt]
2 & add prepositional phrase                                                     & what happened in 1998 ? $\rightarrow$ what happened \textbf{\emph{to Debra Marshall,  Manager of Jeff Jarrett}} in 1998 ?                                     \\ \hdashline[4pt/5pt]

3 & explain *else* context for questions with the forms, e.g., else/other/as well & Was there \textbf{\emph{any other}} views he had in regards to them ? $\rightarrow$ \textbf{\emph{Other than Peter Tatchell condemned the Soviet Union's invasions of Czechoslovakia}}, was there any other views Peter Tatchell had in regards to Soviet Union ? \\ \hdashline[4pt/5pt]
4 & extend the nominal phrase, e.g., name/entity                                  & Who wrote the \textbf{\emph{song}} ? $\rightarrow$ Who wrote the \textbf{\emph{'03 Bonnie \& Clyde song}} ?                                                                                                                                                                \\ \hdashline[4pt/5pt]

5 & expand the special Wh* questions, e.g., why?/what happened/which              & \textbf{\emph{Which}} of the show is the biggest ? $\rightarrow$ \textbf{\emph{Which espisode of The Oprah Winfrey Show}} is the biggest?                                                                                                                                                                    \\ \hdashline[4pt/5pt]

6 & add completed sentences after that/this                                      & What was the aftermath of \textbf{\emph{that}} ? $\rightarrow$ What was the aftermath of \textbf{\emph{Robert Kennedy was chosen by McCarthy as a counsel for ...}} ?                                                                                                                \\ \hdashline[4pt/5pt]

7 & other options                                                                &                                                                                                                              \\ \bottomrule
\end{tabular}%
}
\caption{The commonly used rewriting rules for QR in \textsc{canard}.}
\label{tab:rewrite_rule}
\end{table*}

\subsection{Setup}

\noindent{\textbf{Model Settings.}}\ \ We build our models on the pre-trained language model of BART~\cite{DBLP:conf/acl/LewisLGGMLSZ20}. 
Specifically, we use BART-base to initialize our models. There are 6 transformer layers for the encoder and decoder in BART-base. For our adapter, we map the dimension of the input hidden vector from 768 to 384 which is re-mapped to 768 for the output vector. The hidden vector size for adapter tuning is the default value of 384. Based on BART-base, we need a total of ${\tiny {6 \times 2 + 6 \times 3 = 30}}$ adapters for each private model. We set $\gamma$ to 0.5 in Eq.~\ref{eq:distill} for \textsc{canard} and 0.9 for \textsc{QReCC}. $\alpha$ from Eq.~\ref{eq:alpha} is set to 2 for both \textsc{canard} and \textsc{QReCC}. When fine-tuning BART, we set the learning rate to 1e-5, and for adapter tuning, the learning rate is 1e-4 (both values are tuned from \{1e-4, 1e-5\}). 
We use the validation set to keep the best model based on the BLEU score. We implement our models with HuggingFace~\cite{DBLP:journals/corr/abs-1910-03771} and keep the other default training settings. 
In \textsc{canard}, about 20\% of the questions can be rewritten by replacing pronouns with their referents, so we carry out pronoun replacement first for the questions~(if any) before using BLEU scores to measure rewriting difficulties. More details are given in Appendix \ref{sec:experiment}.

\noindent{\textbf{Baselines.}} \ \ We compare to the following baselines. \textbf{S} denotes training only one shared model with all the training data, which is commonly used in previous work~\cite{DBLP:conf/emnlp/ElgoharyPB19,DBLP:journals/corr/abs-2004-01909}. By adapting BART, \textbf{P-hard}, \textbf{P-medium}, and \textbf{P-easy} are the baselines that train private models on the hard, medium, and easy classes respectively, using fine-tuning or adapter-tuning. Assuming that rewriting difficulty labels are accessible for questions at test time (i.e., the oracle setting), \textbf{Mix-gold} processes a question by the corresponding private model using the difficulty label. \textbf{SLAF} and \textbf{SLAD} denote sequence-level adapter fusion and adapter distillation respectively for combining the private models of P-hard, P-medium, and P-easy. \textbf{SLAF-uni.} combines the private models with uniform distributions. \textbf{SLAF-pred} predicts the class label for the input and then chooses the corresponding private model for generation. 
\textbf{LSTM-S} trains one model using an LSTM-based Seq2Seq model with copy mechanism~\cite{DBLP:conf/acl/SeeLM17} which was used in~\citet{DBLP:conf/emnlp/ElgoharyPB19}.

\noindent{\textbf{Evaluation Metric.}} \ \ Following \citet{DBLP:conf/emnlp/ElgoharyPB19}, we use BLEU\footnote{https://github.com/mjpost/sacrebleu} to obtain the results on hard, medium, and easy classes, and the three results are averaged to obtain the mean result. %We further use ROUGE-1, ROUGE-2, 

\subsection{Robustness Evaluation}\label{sec:robustness-eval}
\subsubsection{Rewriting Difficulty}
We first study rewriting difficulties across different questions. 
Table~\ref{tab:result_3_domains} shows the results on hard, medium, and easy classes on \textsc{canard}. 
\textbf{Each class vs. Overall}: Comparing to the overall results, the rewriting performances of hard questions drop substantially, but are much higher on the easy class. \textbf{LSTM-S vs. BART-S}: By comparing LSTM-S to tuning on BART, LSTM-S achieves higher performance on the easy class but much worse performance on hard and medium classes. This is probably because for easy questions, the model only needs to copy some words from the context and LSTM-S has an explicit copy mechanism to achieve this goal but not BART. Since BART learns a more complex model than LSTM-S, it can better deal with harder questions.  

We further divide the test set into ten classes in Figure~\ref{fig:10_dis}, where the interval $[0, 1]$ is equally divided into ten sub-intervals of size 0.1. We find that when $\mathrm{z}$ gets smaller, rewriting performance degrades, indicating an increase in rewriting difficulty.

\subsubsection{Human Evaluation}\label{sec:human-eval}
The above evaluation results show that our method can effectively divide the questions into subsets with different rewriting difficulties. Here, we conduct a human evaluation to evaluate the question characteristics on these subsets for validity and see what makes the questions hard or easy to rewrite.  

\noindent{\textbf{Question Annotation.}} \ \ To find out what makes the questions different, we first summarize the commonly used rewriting rules, which describe the operations of translating a question into its rewrite. 6 rules are summarized from the training set of \textsc{canard} and presented in Table~\ref{tab:rewrite_rule}. Different rules account for different rewriting hardness for QR systems. For example, the rule of \emph{replace pronoun} is very simple since it only requires the model to determine the pronoun to replace. However, rules 5 and 6 shown in the table will be much harder because the model needs to understand the conversational history well, and the information to be filled in is substantial.

Then we randomly select 50 examples from each subset~(hard, medium, and easy) from the test set and annotate what rules in Table~\ref{tab:rewrite_rule} are used for each example. One question may have multiple rewriting rules. More details are in Appendix \ref{sec:human-assesment}.

\noindent{\textbf{Results.}} \ \ We sum the number of each rewriting rule in each subset and show the distributions of rewriting rules for each subset in Figure~\ref{fig:3_type_dis}. The three distributions are quite different. We find that:
\begin{compactitem}
    \item the easy subset mainly uses rule 1 for rewriting questions;
    \item for medium and hard subsets, other rules are used, such as rules 2, 3, and 4 which are more complex than rule 1;
    \item the hard class uses more rules 2, 3, 5, and 6 compared to the medium class, which demonstrates that the hard class is more difficult than the medium class. 
\end{compactitem}

\noindent{\textbf{Discussion.}} \ \ By knowing the characteristics of each class of questions, we can optimize the model architecture of private models accordingly. For hard questions, we can add some rules to deal with \emph{Wh*} questions. For easy questions, LSTM-based models seem to be good enough as Table~\ref{tab:result_3_domains} indicates. In this work, we have shown that the questions vary in rewriting difficulties and to improve the overall rewriting performance, we focus on the ensemble method to combine the private models. We leave optimizing the model architecture to future work.

% \usepackage{graphicx}
\begin{table}[]
\centering
\resizebox{\columnwidth}{!}{%
\begin{tabular}{lcccc}
\toprule[1.1pt]
\multicolumn{1}{l}{Model $\diagdown$ $\mathcal{D}$}              & \multicolumn{1}{c}{\textbf{Hard}}  & \multicolumn{1}{c}{\textbf{Medium}} & \multicolumn{1}{c}{\textbf{Easy}}  & \textbf{Mean}  \\ \hline
\multicolumn{5}{l}{\underline{\emph{LSTM based}}}                                                                                                                                    \\ 
\multicolumn{1}{l}{S}             & \multicolumn{1}{c}{$26.29$}        & \multicolumn{1}{c}{$50.79$}         & \multicolumn{1}{c}{$79.41$}        & $52.16$        \\ 
\multicolumn{1}{l}{Mix-gold}      & \multicolumn{1}{c}{$27.79$}        & \multicolumn{1}{c}{$51.91$}         & \multicolumn{1}{c}{$86.53$}        & $55.41$        \\ \hline
\multicolumn{5}{l}{\underline{\emph{Fine-tune BART-base}}}                                                                                                                               \\ 
\multicolumn{1}{l}{S}             & \multicolumn{1}{c}{$39.38$}        & \multicolumn{1}{c}{$53.7$}          & \multicolumn{1}{c}{$66.33$}        & $53.14$        \\
\multicolumn{1}{l}{Mix-gold}      & \multicolumn{1}{c}{$40.91$}        & \multicolumn{1}{c}{$56.15$}         & \multicolumn{1}{c}{$74.00$}        & $57.02$        \\ \hline
\multicolumn{5}{l}{\underline{\emph{Tuning BART-base with adapters}}}                                                         \\ 
\multicolumn{1}{l}{S}             & \multicolumn{1}{c}{$39.20_{0.52}$} & \multicolumn{1}{c}{$53.14_{0.11}$}  & \multicolumn{1}{c}{$65.97_{0.12}$} & $52.77_{0.16}$ \\ 
\multicolumn{1}{l}{P-hard}        & \multicolumn{1}{c}{$41.33_{0.27}$} & \multicolumn{1}{c}{$46.39_{0.46}$}  & \multicolumn{1}{c}{$55.24_{0.93}$} & $47.66_{0.51}$ \\ 
\multicolumn{1}{l}{P-medium}      & \multicolumn{1}{c}{$34.41_{0.19}$} & \multicolumn{1}{c}{$54.68_{0.31}$}  & \multicolumn{1}{c}{$62.98_{0.14}$} & $50.69_{0.11}$ \\ 
\multicolumn{1}{l}{P-easy}        & \multicolumn{1}{c}{$27.42_{0.26}$} & \multicolumn{1}{c}{$55.55_{0.16}$}  & \multicolumn{1}{c}{$73.63_{0.18}$} & $52.20_{0.12}$ \\ 
\multicolumn{1}{l}{SLAF-uni.}     & \multicolumn{1}{c}{$34.05_{0.09}$} & \multicolumn{1}{c}{$55.88_{0.65}$}  & \multicolumn{1}{c}{$67.27_{0.09}$} & $52.40_{0.23}$ \\ 
\multicolumn{1}{l}{SLAF-pred}     & \multicolumn{1}{c}{$32.96_{0.26}$} & \multicolumn{1}{c}{$55.62_{0.38}$}  & \multicolumn{1}{c}{$70.83_{0.21}$} & $53.14_{0.12}$ \\
\multicolumn{1}{l}{\textbf{SLAF}} & \multicolumn{1}{c}{$34.55_{0.05}$} & \multicolumn{1}{c}{$56.05_{0.32}$}  & \multicolumn{1}{c}{$69.05_{0.15}$} & $^*53.22_{0.17}$ \\ 
\multicolumn{1}{l}{\textbf{SLAD}} & \multicolumn{1}{c}{$38.26_{0.39}$} & \multicolumn{1}{c}{$54.22_{0.10}$}  & \multicolumn{1}{c}{$67.57_{0.30}$} & $^*53.35_{0.17}$ \\
\multicolumn{1}{l}{Mix-gold}      & \multicolumn{1}{c}{$41.33_{0.27}$} & \multicolumn{1}{c}{$54.68_{0.31}$}  & \multicolumn{1}{c}{$73.63_{0.18}$} & $56.55_{0.12}$ \\ \bottomrule[1.1pt]
\end{tabular}%
}
\caption{The test results~(mean and standard deviation) on \textsc{canard}. We run 3 times for adapter tuning. $*$ indicates statistically significant improvement over S and  SLAF-uni.~($p < 0.05$).}
\label{tab:main-result-on-cannard}
\end{table}

% \usepackage{graphicx}
\begin{table}[]
\centering
\resizebox{\columnwidth}{!}{%
\begin{tabular}{lcccc}
\toprule[1.1pt]
 \multicolumn{1}{l}{Model $\diagdown$ $\mathcal{D}$} & \multicolumn{1}{c}{\textbf{Hard}} & \multicolumn{1}{c}{\textbf{Medium}} & \multicolumn{1}{c}{\textbf{Easy}} & \multicolumn{1}{c}{\textbf{Mean}} \\ \hline
 \multicolumn{5}{l}{\underline{\emph{Tuning BART-base with adapters}}}  \\
S     & $45.43_{0.27}$ & $60.60_{0.21}$ & $78.47_{0.17}$ & $61.50_{0.02}$  \\ 
P-hard        & $49.48_{0.16}$ & $53.13_{0.09}$ & $67.32_{0.19}$ & $56.65_{0.10}$ \\ 
P-medium      & $43.17_{0.28}$ & $61.83_{0.26}$ & $76.63_{1.19}$ & $60.54_{0.50}$ \\ 
P-easy        & $37.56_{0.80}$ & $63.17_{0.19}$ & $82.79_{0.40}$ & $61.17_{0.22}$ \\ 
SLAF-uni.     & $43.28_{0.39}$ & $62.21_{0.23}$ & $78.89_{0.17}$ & $61.46_{0.17}$ \\ 
SLAF-pred     & $43.60_{0.72}$ & $61.69_{0.64}$ & $79.05_{0.92}$ & $61.45_{0.28}$ \\ 
\textbf{SLAF} & $43.76_{0.53}$ & $62.13_{0.19}$ & $79.71_{0.24}$ & $^*61.87_{0.17}$ \\ 
\textbf{SLAD} & $44.99_{0.25}$ & $61.35_{0.21}$ & $79.93_{0.08}$ & $^*62.09_{0.05}$ \\ 
Mix-gold      & $49.48_{0.16}$ & $61.83_{0.26}$ & $82.79_{0.40}$ & $64.70_{0.09}$ \\ \bottomrule[1.1pt]
\end{tabular}%
}
\caption{The test results~(mean and standard deviation) on \textsc{QReCC}. We run 3 times for adapter tuning. $*$ indicates statistically significant improvement over S, SLAF-uni., and SLAF-pred~($p < 0.05$).}
\label{tab:main-result-on-qrecc}
\end{table}

\subsection{Question Rewriting}
We report our results on question rewriting based on \textsc{canard} and \textsc{QReCC}. From the results in Tables~\ref{tab:main-result-on-cannard} and \ref{tab:main-result-on-qrecc}, we first show the results of each class, then the mean performances are displayed. 
\textbf{Mix-gold, SLAF, SLAD  vs. S}: (\textbf{a}) Mix-gold, SLAF, and SLAD are consistently better than S, which demonstrates the effectiveness of learning private models to model multiple underlying distributions. (\textbf{b}) From the results on each class, SLAF and SLAD can substantially enhance the performance on medium and easy classes compared to S. (\textbf{c}) SLAD is more effective than SLAF and SLAD is more efficient during inference. (\textbf{d}) We find Mix-gold to be better than SLAF and SLAD, since Mix-gold is an oracle model that uses the correct difficulty label to select the private model for inference. 

We find that by learning a private model for each class, the performance on the corresponding class can be consistently improved, which explains why Mix-gold, SLAF, and SLAD can outperform S. We also find that the sole private model cannot improve the overall rewriting performance of the three classes, but SLAF and SLAD can outperform S after model ensemble, which demonstrates the necessity of combining the private models.

\noindent{\textbf{Model Ensemble.}} \ \ One question is whether the improvements of SLAF and SLAD simply come from combing multiple models and whether applying only one private model selected by the predicted class label is better. As shown in Tables~\ref{tab:main-result-on-cannard} and ~\ref{tab:main-result-on-qrecc}, we find SLAF-uni. performs worse than SLAF and SLAD, which demonstrates that the benefits of SLAF and SLAD are not simply because of the model ensemble, but class estimation also helps~(In SLAD, class estimation lies in using gold class labels of questions for knowledge distillation during training). SLAF-pred can be regarded as an ensemble method since it uses multiple private models during inference. Compared to SLAF, SLAF-pred uses one-hot class weights to combine the private models. However, SLAF-pred performs worse than SLAF, and the reason could be that classifying the question into the corresponding class is nontrivial, wrong predictions will have much worse rewriting results as the results of P-hard, -medium, -easy on other classes indicate.

\subsection{Further Analysis}\label{sec:FA}

% \usepackage{graphicx}
\begin{table}[]
\centering
\small
\resizebox{\columnwidth}{!}{%
\begin{tabular}{lccccc}
\toprule[.8pt]
Method $\diagdown$ $\mathcal{D}$ & $\mathcal{D}_1$ & $\mathcal{D}_2$ & $\mathcal{D}_3$ & Trend & Std.  \\ \hline
$|\mathrm{q}|$               & 5.27            & 7.22            & 10.04    & $\nearrow$ & $-$     \\ 
BLEU      & 38.90           & 52.78           & 63.29    &  $\nearrow$ & 12.2      \\ \hline
$|\mathrm{q'}|$              & 6.82            & 10.14           & 17.07    & $\nearrow$ &    $-$   \\ 
BLEU                         & 39.46           & 61.51           & 50.08    & $\nearrow$, $\searrow$ & 11.0       \\ \hline
$|\mathrm{q}|/|\mathrm{q'}|$ & 0.47            & 0.75            & 0.97     & $\nearrow$  &  $-$   \\ 
BLEU                         & 43.18           & 55.93           & 56.69    & $\nearrow$    & 7.6   \\ \hline
ROUGE-L~(\%)                 & 56.25           & 76.04           & 94.25    & $\nearrow$    & $-$    \\ 
BLEU                         & 40.79           & 50.39           & 74.20     & $\nearrow$  & 17.2     \\ \hline
BLEU~(\%)                         & 16.13           & 44.59           & 90.27  & $\nearrow$     & $-$     \\ 
BLEU                        & 39.58           & 53.80           & 65.60    & $\nearrow$    & 13.0     \\ \bottomrule[.8pt]
\end{tabular}%
}
\caption{Results of measuring rewriting difficulty on \textsc{canard}.}
\label{tab:division_analysis}
\end{table}

\noindent{\textbf{Analysis of Rewriting Difficulty Measures.}} \ \ In our work, we use BLEU to measure the discrepancy between a question and its rewrite. We further experiment with other methods to assess their effectiveness for difficulty measurement. \textsc{canard} is evaluated here. As shown in Table~\ref{tab:division_analysis}, we first use the length of a question~($|\mathrm{q}|$), its rewrite~($|\mathrm{q}'|$), and their ratio~($|\mathrm{q}|/|\mathrm{q}'|$) to calculate a difficulty score. After re-ranking the questions with a difficulty score, we divide the ranked questions equally into three classes. Interestingly, we find that $|\mathrm{q}|$ works well. After analysis, we find that rewriting short questions requires finding much missing information, which makes short questions hard questions. The $|\mathrm{q}|/|\mathrm{q}'|$ metric is not very useful, since $|\mathrm{q}|/|\mathrm{q}'|$ can only measure the discrepancy in question lengths, but does not necessarily measure their semantic difference. $|\mathrm{q}'|$ does not work for difficulty measurement. Not surprisingly, the ROUGE score is also useful in measuring discrepancy just like BLEU.

% \usepackage{graphicx}
\begin{table}[]
\centering
\resizebox{\columnwidth}{!}{%
\begin{tabular}{c|ccccccccccc|}
   & 0    & 1    & 2    & 3    & 4    & 5    & 6    & 7    & 8    & 9    & 10   \\ \hline
0  & \cellcolor{gray!60}{19.2} & \cellcolor{gray!54.6}{28.3} & \cellcolor{gray!41}{34.7} & \cellcolor{gray!19.1}{39.9} & \cellcolor{gray!13.7}{44.2} & \cellcolor{gray!9.4}{50.3} &  \cellcolor{gray!8.8}{57.9} & \cellcolor{gray!8.6}{64.6} & \cellcolor{gray!12.2}{71.6} & \cellcolor{gray!9.6}{80.3} & \cellcolor{gray!6.4}{71.9} \\
1  & \cellcolor{gray!28.9}{17.7} & \cellcolor{gray!51.2}{28.1} & \cellcolor{gray!58.5}{36.1} & \cellcolor{gray!40}{43.3} & \cellcolor{gray!31.5}{48.5} & \cellcolor{gray!16.7}{53.6} & \cellcolor{gray!14.9}{61.4} & \cellcolor{gray!11.1}{66.5} & \cellcolor{gray!17.2}{74.5} & \cellcolor{gray!5.3}{75.1} & \cellcolor{gray!8.8}{74.5} \\
2  & \cellcolor{gray!11.6}{16.0} & \cellcolor{gray!60}{28.6} & \cellcolor{gray!60}{36.2} & \cellcolor{gray!46.2}{44.0} & \cellcolor{gray!36.5}{49.3} & \cellcolor{gray!24.3}{55.9} & \cellcolor{gray!23.8}{64.7} & \cellcolor{gray!18.3}{70.3} & \cellcolor{gray!31.2}{79.6} & \cellcolor{gray!18.2}{86.2} & \cellcolor{gray!14.3}{78.6} \\
3  & \cellcolor{gray!6.5}{15.0} & \cellcolor{gray!33.4}{26.8} & \cellcolor{gray!52.9}{35.7} & \cellcolor{gray!60}{45.3} & \cellcolor{gray!52.2}{51.3} & \cellcolor{gray!31.4}{57.5} & \cellcolor{gray!32.2}{66.9} & \cellcolor{gray!19.5}{70.8} & \cellcolor{gray!32.6}{80.0} & \cellcolor{gray!22.9}{88.4} & \cellcolor{gray!19.1}{81.2} \\
4  & \cellcolor{gray!1.6}{12.8} & \cellcolor{gray!25.4}{26.0} & \cellcolor{gray!55.7}{35.9} & \cellcolor{gray!54.3}{44.8} & \cellcolor{gray!60}{52.1} & \cellcolor{gray!46.7}{60.1} & \cellcolor{gray!41.9}{68.9} & \cellcolor{gray!27.4}{73.5} & \cellcolor{gray!27.5}{78.5} & \cellcolor{gray!46.7}{95.7} & \cellcolor{gray!20.4}{81.8} \\
5  & \cellcolor{gray!1.3}{12.5} & \cellcolor{gray!19.9}{25.3} & \cellcolor{gray!45.5}{35.1} & \cellcolor{gray!55.4}{44.9} & \cellcolor{gray!43.7}{50.3} & \cellcolor{gray!54.2}{61.1} & \cellcolor{gray!50.9}{70.4} & \cellcolor{gray!36.5}{75.9} & \cellcolor{gray!32.3}{79.9} & \cellcolor{gray!39.8}{94.0} & \cellcolor{gray!27.1}{84.4} \\
6  & \cellcolor{gray!0.8}{11.8} & \cellcolor{gray!17.9}{25.0} & \cellcolor{gray!43.2}{34.9} & \cellcolor{gray!50.1}{44.4} & \cellcolor{gray!56}{51.7} & \cellcolor{gray!59.1}{61.7} & \cellcolor{gray!54.9}{71.0} & \cellcolor{gray!43.6}{77.4} & \cellcolor{gray!40.3}{81.9} & \cellcolor{gray!25.3}{89.4} & \cellcolor{gray!34.5}{86.7} \\
7  & \cellcolor{gray!0.8}{11.9} & \cellcolor{gray!14.4}{24.4} & \cellcolor{gray!38.9}{34.5} & \cellcolor{gray!48.1}{44.2} & \cellcolor{gray!54.1}{51.5} & \cellcolor{gray!60}{61.8} & \cellcolor{gray!60}{71.7} & \cellcolor{gray!60}{80.2} & \cellcolor{gray!55.7}{84.9} & \cellcolor{gray!30}{91.1} & \cellcolor{gray!39}{87.9} \\
8  & \cellcolor{gray!0.1}{9.4}  & \cellcolor{gray!3.4}{20.8} & \cellcolor{gray!16.2}{31.3} & \cellcolor{gray!28.5}{41.7} & \cellcolor{gray!37.2}{49.4} & \cellcolor{gray!37.2}{58.6} & \cellcolor{gray!37.7}{68.1} & \cellcolor{gray!37}{76.0} & \cellcolor{gray!60}{85.6} & \cellcolor{gray!55.7}{97.6} & \cellcolor{gray!58.8}{92.0} \\
9  & \cellcolor{gray!10.4}{15.8} & \cellcolor{gray!39.5}{27.3} & \cellcolor{gray!47.8}{35.3} & \cellcolor{gray!53.2}{44.7} & \cellcolor{gray!48.6}{50.9} & \cellcolor{gray!47.4}{60.2} & \cellcolor{gray!45.3}{69.5} & \cellcolor{gray!35.3}{75.6} & \cellcolor{gray!49}{83.7} & \cellcolor{gray!25.3}{89.4} & \cellcolor{gray!31.7}{85.9} \\
10 & \cellcolor{gray!2.5}{13.5} & \cellcolor{gray!16}{24.7} & \cellcolor{gray!42.1}{34.8} & \cellcolor{gray!50.1}{44.4} & \cellcolor{gray!58}{51.9} & \cellcolor{gray!47.4}{60.2} & \cellcolor{gray!46.5}{69.7} & \cellcolor{gray!34.4}{75.4} & \cellcolor{gray!40.8}{82.0} & \cellcolor{gray!60}{98.4} & \cellcolor{gray!60}{92.2} \\ \hline
\end{tabular}%
}
\caption{BLEU scores for different classes on \textsc{canard}. The rows are the private models and the columns are the classes.}
\label{tab:heatmap}
\end{table}

\noindent{\textbf{Analysis of Learning Data Distribution.}} \ \ Tables~\ref{tab:main-result-on-cannard} and \ref{tab:main-result-on-qrecc} show that learning private models can enhance performance on each class. We further divide the data into eleven classes~($\mathrm{z} \in$ $[0,0.1]$, $(0.1,0.2]$, $\cdots$, $(0.9,1)$, $1$) and learn a private model for each class. We build the private models using LSTM-S, in which we first train a shared model on the full training data, then fine-tune the shared model on each class to obtain the private models. Table~\ref{tab:heatmap} shows the BLEU scores where the score in the ($i, j$) entry is obtained by training on class $i$ and testing on class $j$. On the whole, learning private models can enhance the performance of the corresponding class. With these private models, we can better model the data distributions, but how to combine a large number of private models is a challenge, since it is hard to train a classifier to correctly predict so many class labels, which will have some negative effects on the model ensemble. 

\begin{figure}[t]
\begin{center}
\includegraphics[width=7cm]{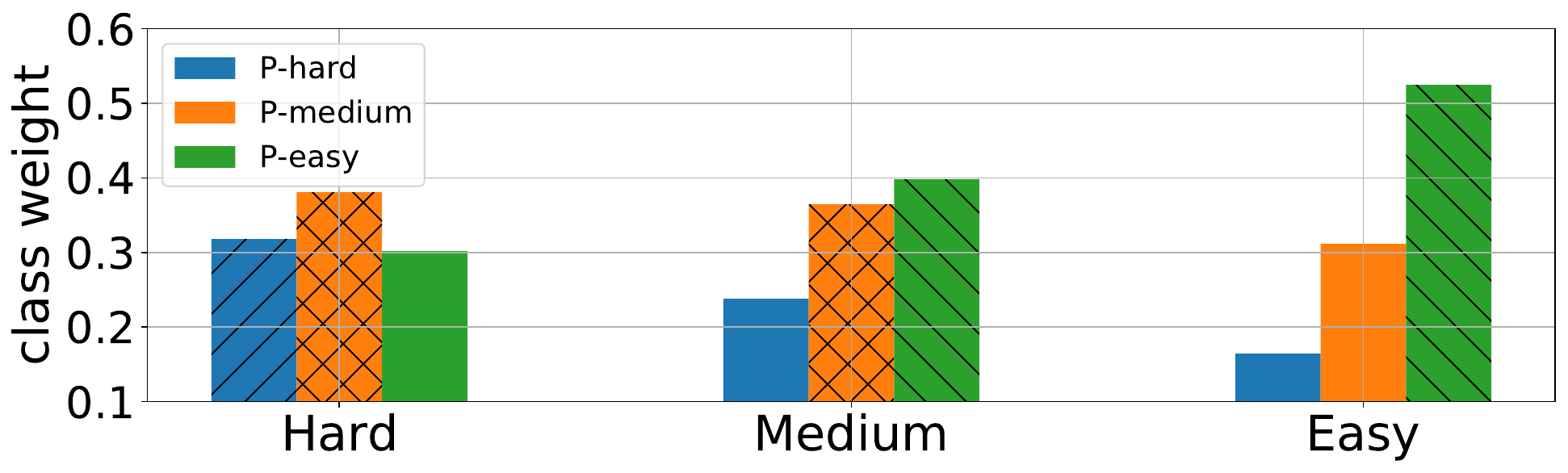}
\end{center}
\caption{Class weights for different classes on \textsc{canard}.}
\label{fig:domain_distribution}
\end{figure}

\begin{figure}[t]
\begin{center}
\includegraphics[width=7cm]{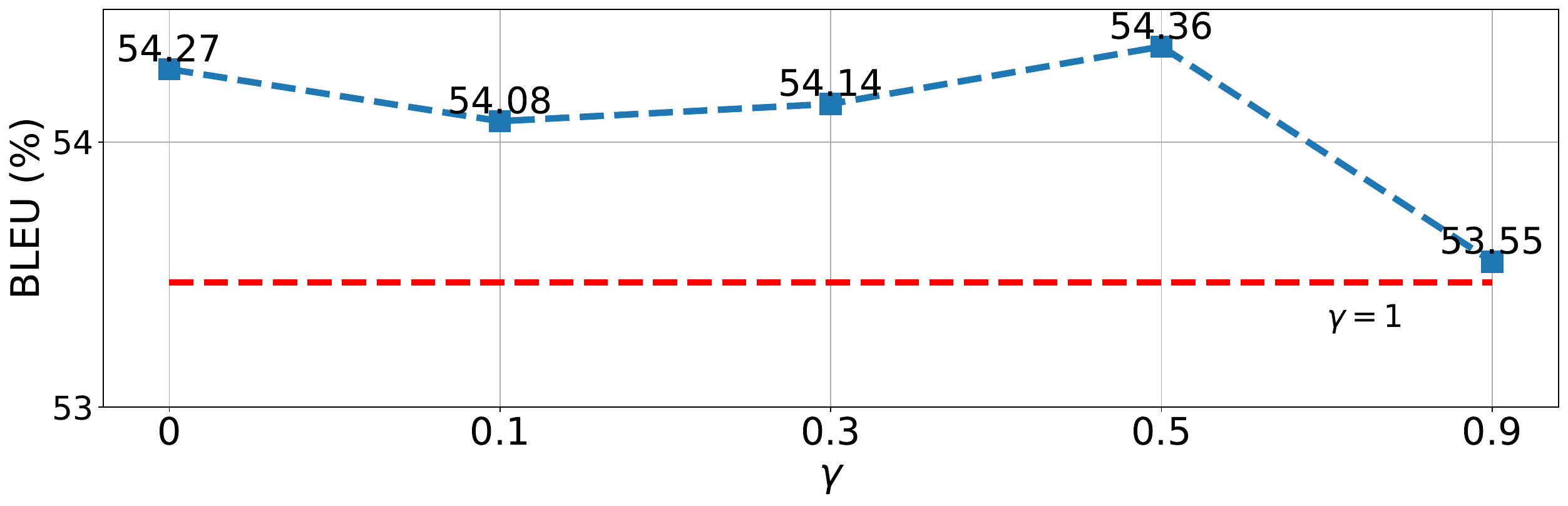}
\end{center}
\caption{BLEU socres for different $\gamma$ values on \textsc{canard}.}
\label{fig:gamma}
\end{figure}

\noindent{\textbf{Analysis of SLAF \& SLAD.}} \ \ We plot the class distributions of hard, medium, and easy classes in Figure~\ref{fig:domain_distribution}. We find that in the hard class, the class weights are almost equally distributed among the private models, which means that the hard questions are difficult for classification. This result explains why SLAF performs worse than S for hard questions in Tables~\ref{tab:main-result-on-cannard} and \ref{tab:main-result-on-qrecc}. 
We further study the contribution of distillation in SLAD. In Figure~\ref{fig:gamma}, on the whole, when $\gamma$ increases, the contribution of distillation decreases, and the performance drops, indicating that distillation is important for SLAD. 

\noindent{\textbf{Case Study.}} \ \ We further show generated rewriting samples of various methods on \textsc{canard} in Appendix \ref{sec:case_study}.

\section{Conclusion}
In this work, we study the robustness of a QR system to questions varying in rewriting hardness. We use a simple yet effective heuristic to measure the rewriting difficulty. We further propose a novel method to deal with varying rewriting difficulties. Tested on \textsc{canard} and \textsc{QReCC}, we show the effectiveness of our methods.

\section*{Acknowledgments}

This research is supported by the National Research Foundation, Singapore under its AI Singapore Programme (AISG Award No: AISG-RP-2018-007 and AISG2-PhD-2021-08-016[T]). The computational work for this article was partially
performed on resources of the National Supercomputing Centre, Singapore (https://www.nscc.sg).

\bibliographystyle{acl_natbib}
\bibliography{anthology,acl2021}

%\appendix

\appendix
%\onecolumn

%\begin{center}
%\Large 
%\textbf{
%Difficulty-Aware Rewriting of Questions in Context\\(Supplementary Materials)}
%\end{center}

\section{Experimental Setup}\label{sec:experiment}
We use HuggingFace~\cite{DBLP:journals/corr/abs-1910-03771} to implement our model. We follow the training script from \url{https://github.com/huggingface/transformers/tree/master/examples/seq2seq} to train the model. Models are trained for 10 epochs. Batch size is selected from \{10, 16, 32\}. Learning rate is selected from \{1e-5, 1e-4\}. We train 10 epochs for \textsc{CANARD} and 8 epochs for \textsc{QReCC}. The best model based on the BLEU score on the validation set is kept. The beam width for beam search is the default value of 4. 

For our QR framework, we first train a private model for each class. For model ensemble, the weights of the private models are frozen without updating. On \textsc{QReCC}, to build the private models, on each class of data, we fine-tune the shared model which is trained on all the training data, since we find that this can enhance the final performance, but on \textsc{canard}, we do not see the improvement. The learning rate of fine-tuning in this process is 1e-5.

To pre-process the dataset, we only tokenize the sentences. And we append the question and its history context with ``$|||$''.

In \textsc{canard}, about 20\% of the questions can be rewritten by replacing pronouns with their referents, so we carry out pronoun replacement first for the questions~(if any) before using BLEU scores to measure rewriting difficulties.

\section{Human Assessment for Rewriting Rules}\label{sec:human-assesment}

We first ask one annotator to summarize some common rewriting rules by looking at the training set of \textsc{canard}~\cite{DBLP:conf/emnlp/ElgoharyPB19}. When accessing the rewriting rules used for each question, the second annotator will rely on the summarized rewriting rules for annotation. For each class, we randomly select 50 questions from the test set for annotation. 

\noindent{\textbf{Case Study.}} \ \ Table~\ref{tab:case_study} shows some annotated results from the hard, medium, and easy classes.

\begin{sidewaystable*}[]
\centering
\resizebox{25cm}{!}{%
\begin{tabular}{p{7cm}|p{7cm}|p{3cm}|p{2cm}|p{2cm}|p{2cm}|p{2cm}|p{2cm}|p{2cm}}
\toprule
                                                             &                                                                                                                                                                                                           & 1                                                         & 2                        & 3                                                                            & 4                                           & 5                                                               & 6                                       & 7             \\ \hline
Question                                                     & Rewrite                                                                                                                                                                                                 & replace pronoun, e.g., he/ his/ she/ her/ they/ their/ it/ its... & add prepositional phrase & explain *else* context for questions with the forms, e.g., else/other/as well & extend the nominal phrase, e.g., name/entity & expand the special Wh* questions, e.g., why?/what happened/which & add completed sentences after that/this & other options \\ \hline
\multicolumn{1}{c}{\textbf{easy}}                            &                                                                                                                                                                                                           &                                                           &                          &                                                                              &                                             &                                                                 &                                         &               \\ 
How does he try to take over the world ?                     & How does Brain try to take over the world ?                                                                                                                                                               & 1                                                         &                          &                                                                              &                                             &                                                                 &                                         &               \\ \hdashline[3pt/5pt]
what were some of his careers that involved flying ?         & what were some of Luis Walter Alvarez 's careers that involved flying ?                                                                                                                                   & 1                                                         &                          &                                                                              &                                             &                                                                 &                                         &               \\ \hdashline[3pt/5pt]
What was so special about the song ?                         & What was so special about the song Purple Haze ?                                                                                                                                                          &                                                           &                          &                                                                              & 1                                           &                                                                 &                                         &               \\ \midrule
\multicolumn{1}{c}{\textbf{medium}}                          &                                                                                                                                                                                                           &                                                           &                          &                                                                              &                                             &                                                                 &                                         &               \\ 
When did she start to campaign ?                             & When did Jeanine Pirro start to campaign for the lieutenant governor role ?                                                                                                                               & 1                                                         & 1                        &                                                                              &                                             &                                                                 &                                         &               \\ \hdashline[3pt/5pt]
Did this lead to another choreographing job ?                & Did Etude in D Minor lead to another choreographing job ?                                                                                                                                                 & 2                                                         &                          &                                                                              &                                             &                                                                 &                                         &               \\ \hdashline[3pt/5pt]
What did they do after the album ?                           & What did Fat Freddy 's Drop do after the album , " Live at the Matterhorn " ?                                                                                                                             & 1                                                         &                          &                                                                              & 1                                           &                                                                 &                                         &               \\ \midrule
\multicolumn{1}{c}{\textbf{hard}}                            &                                                                                                                                                                                                           &                                                           &                          &                                                                              &                                             &                                                                 &                                         &               \\ 
Was there anything else noteworthy about the autobiography ? & Aside from Chester 's friends being uncomfortable with his writing , was there anything else noteworthy about Chester Brown 's autobiography ?                                                            & 1                                                         &                          & 1                                                                            &                                             &                                                                 &                                         &               \\ \hdashline[3pt/5pt]
was the tour domestic or international ?                     & was the Sanctus Real 's Fight the Tide Tour domestic or international ?                                                                                                                                   &                                                           &                          &                                                                              & 1                                           &                                                                 &                                         &               \\ \hdashline[3pt/5pt]
Was there any other views he had in regards to them ?        & Other than Peter Tatchell condemned the Soviet Union 's invasions of Czechoslovakia , Afghanistan and its internal repression , was there any other views Peter Tatchell had in regards to Soviet Union ? & 2                                                         &                          &     1                                                                         & 1                                           &                                                                 &                                         &               \\ \bottomrule
\end{tabular}%
}

\caption{The case study of human annotation} 
\label{tab:case_study}
\end{sidewaystable*}

\section{Case Study of Generated Rewrites}\label{sec:case_study}

We further show some cases of generated rewrites from various methods (S, Mix-gold, SLAF, and SLAD). We use adapter tuning to build these models. Tables~\ref{tab:hard}, \ref{tab:medium}, and \ref{tab:easy} show the generated rewrites on hard, medium, and easy classes respectively.

% \usepackage{graphicx}
\begin{table*}[]
\centering
\resizebox{\textwidth}{!}{%
\begin{tabular}{lp{14cm}}
\toprule
\multicolumn{2}{c}{\textbf{Hard}}                                                                                                                                                          \\ 
\textbf{Models} & \textbf{Generated Rewrites}                                                                                                                                            \\ \hline
Reference       & In addition to Ezio Pinza 's role in La Vestale and his performance of Don Giovanni , are there any other interesting aspects about this article ?                       \\  \hdashline[3pt/5pt]
S               & Besides Ezio Pinza 's operas , are there any other interesting aspects about this article ?                                                                              \\  \hdashline[3pt/5pt]
Mix-gold        & Besides Ezio Pinza singing Don Giovanni in Spontini 's La vestale , are there any other interesting aspects about this article ?                                         \\  \hdashline[3pt/5pt]
SLAF             & Are there any other interesting aspects about this article besides Ezio Pinza 's operas ?                                                                                \\  \hdashline[3pt/5pt]
SLAD             & Are there any other interesting aspects about this article besides Ezio Pinza 's operas ?                                                                                \\ \midrule
Reference       & did the scathing review by Saibal Chatterjee have a bad effect on Kapoor 's future work ?                                                                                \\  \hdashline[3pt/5pt]
S               & did Phata Poster Nikhla Hero have a bad effect on Shahid Kapoor 's future work ?                                                                                         \\  \hdashline[3pt/5pt]
Mix-gold        & Did the negative review of Phata Poster Nikhla Hero have a bad effect on Shahid Kapoor 's future work ?                                                                  \\  \hdashline[3pt/5pt]
SLAF             & did the comedy Phata Poster Nikhla Hero have a bad effect on Shahid Kapoor 's future work ?                                                                              \\  \hdashline[3pt/5pt]
SLAD             & Did Phata Poster Nikhla Hero have a bad effect on Shahid Kapoor 's future work ?                                                                                         \\ \midrule
Reference       & Besides his college honors , what other awards did Victor Davis Hanson win ?                                                                                             \\  \hdashline[3pt/5pt]
S               & What other awards did Victor Davis Hanson win besides his BA and PhD ?                                                                                                   \\  \hdashline[3pt/5pt]
Mix-gold        & Besides being a Senior Fellow at the Hoover Institution and professor emeritus at California State University , Fresno , what other awards did Victor Davis Hanson win ? \\  \hdashline[3pt/5pt]
SLAF             & Besides the awards , what other awards did Victor Davis Hanson win ?                                                                                                     \\  \hdashline[3pt/5pt]
SLAD             & What other awards did Victor Davis Hanson win other than being a Senior Fellow at the Hoover Institution and professor emeritus at California State University ?         \\ \bottomrule
\end{tabular}%
}
\caption{Generated rewrites on the hard class.}
\label{tab:hard}
\end{table*}

% \usepackage{graphicx}
\begin{table*}[]
\centering
\resizebox{\textwidth}{!}{%
\begin{tabular}{lp{14cm}}
\toprule
\multicolumn{2}{c}{\textbf{Medium}}                                                                                                                                                                                      \\ 
\textbf{Models} & \textbf{Generated Rewrites}                                                                                                                                                                          \\ \hline
Reference       & Besides Do or Die any other chart toppers ?                                                                                                                                                            \\  \hdashline[3pt/5pt]
S               & Besides Do or Die , did Super Furry Animals have any other chart toppers ?                                                                                                                             \\  \hdashline[3pt/5pt]
Mix-gold        & Besides Do or Die , any other chart toppers ?                                                                                                                                                          \\  \hdashline[3pt/5pt]
SLAF             & Besides Guerrilla , any other chart toppers ?                                                                                                                                                          \\  \hdashline[3pt/5pt]
SLAD             & Besides ” Do or Die ” , any other chart toppers ?                                                                                                                                                      \\ \midrule
Reference       & What did Jeanine Pirro do after running for lieutenant governor ?                                                                                                                                      \\  \hdashline[3pt/5pt]
S               & What did Jeanine Pirro do next after being Assistant District Attorney of Westchester County ?                                                                                                         \\ \hdashline[3pt/5pt]
Mix-gold        & What did Jeanine Pirro do after writing appeals ?                                                                                                                                                      \\ \hdashline[3pt/5pt]
SLAF             & What did Jeanine Pirro do next after serving as Assistant District Attorney ?                                                                                                                          \\  \hdashline[3pt/5pt]
SLAD             & What did Jeanine Pirro do next after being appointed Assistant District Attorney of Westchester County ?                                                                                               \\ \midrule
Reference       & Besides trouble adapting to the faster pace of the Premiership , what else was bad about Juan Sebastián Verón 's time at Old Trafford ?                                                                \\  \hdashline[3pt/5pt]
S               & Besides having trouble adapting to the faster pace of the Premiership and being not allowed the same space and time on the ball , what else was bad about Juan Sebastián Verón 's time at Old Trafforr \\  \hdashline[3pt/5pt]
Mix-gold        & What else was bad about Juan Sebastián Verón 's time at Old Trafford besides adapting to the faster pace of the Premiership ?                                                                          \\  \hdashline[3pt/5pt]
SLAF             & What else was bad about Juan Sebastián Verón 's time at Old Trafford other than the faster pace of the Premiership ?                                                                                   \\  \hdashline[3pt/5pt]
SLAD             & What else was bad about Juan Sebastián Verón 's time at Old Trafford other than being not allowed the same space and time on the ball ?                                                                \\ \bottomrule
\end{tabular}%
}
\caption{Generated rewrites on the medium class.}
\label{tab:medium}
\end{table*}

% \usepackage{graphicx}
\begin{table*}[]
\centering
\resizebox{\textwidth}{!}{%
\begin{tabular}{lp{14cm}}
\toprule
\multicolumn{2}{c}{\textbf{Easy}}                                                                                                                     \\ 
\textbf{Models} & \textbf{Generated Rewrites}                                                                                                       \\ \hline
Reference       & Did Robert Fripp win any awards ?                                                                                                   \\  \hdashline[3pt/5pt]
S               & Did Robert Fripp win any awards for his music ?                                                                                     \\  \hdashline[3pt/5pt]
Mix-gold        & Did Robert Fripp win any awards ?                                                                                                   \\  \hdashline[3pt/5pt]
SLAF             & Did Robert Fripp win any awards for Biography ?                                                                                     \\  \hdashline[3pt/5pt]
SLAD             & Did Robert Fripp win any awards ?                                                                                                   \\ \midrule
Reference       & Are there any other interesting aspects about this article aside from Brown collaborating ?                                         \\  \hdashline[3pt/5pt]
S               & Besides Chester Brown bringing Ed to an abrupt end in Yummy Fur \# 18 , are there any other interesting aspects about this article ? \\  \hdashline[3pt/5pt]
Mix-gold        & Besides Ed , Are there any other interesting aspects about this article ?                                                           \\  \hdashline[3pt/5pt]
SLAF             & Besides Chester Brown bringing Ed to an abrupt end in Yummy Fur \# 18 , are there any other interesting aspects about this article ? \\  \hdashline[3pt/5pt]
SLAD             & Besides Chester Brown bringing Ed to an abrupt end in Yummy Fur \# 18 , are there any other interesting aspects about this article ? \\ \midrule
Reference       & What are some of the Green Day related works ?                                                                                      \\  \hdashline[3pt/5pt]
S               & What are some of Jason White 's Green Day related works ?                                                                           \\  \hdashline[3pt/5pt]
Mix-gold        & What are some of the Green Day related works ?                                                                                      \\  \hdashline[3pt/5pt]
SLAF             & What are some of the Green Day related works ?                                                                                      \\  \hdashline[3pt/5pt]
SLAD             & What are some of Jason White 's Green Day related works ?                                                                           \\ \bottomrule
\end{tabular}%
}
\caption{Generated rewrites on the easy class.}
\label{tab:easy}
\end{table*}

%\newpage
%\bibliographystyle{acl_natbib}
%\bibliography{anthology,acl2021}

\end{document}